# SELF-SUPERVISED PERCEPTUALLY INTERPRETABLE MONOCULAR DEPTH ESTIMATION

*Zain Ul Abidin, George Dimas, and Dimitris K. Iakovidis*

Department of Computer Science and Biomedical Informatics, School of Science, University of Thessaly, Lamia, Greece

## ABSTRACT

Self-supervised monocular depth estimation (MDE) enables depth prediction from monocular images without requiring ground-truth supervision, making it attractive for large-scale and real-world applications. Despite steady improvements in accuracy, most existing methods remain difficult to interpret, as depth is inferred from RGB representations that obscure the impact of individual perceptual image components. This lack of transparency limits systematic analysis of failure cases and reduces confidence in safety-critical settings. This paper presents a self-supervised framework for perceptually interpretable monocular depth estimation (PIMDE), designed to associate depth predictions with distinct perceptual components of the input image. Rather than operating directly on RGB inputs, the proposed method decomposes each image into a set of perceptual feature maps (PFMs), each encoding a specific visual cue. Distinct depth estimation branches process these PFMs independently to produce depth estimates (PIDEs), which are subsequently combined through an explicit fusion strategy. This formulation allows us to examine directly the contribution of each perceptual cue to the final depth prediction. Experiments conducted on the KITTI benchmark dataset demonstrate that PIMDE achieves performance comparable to established self-supervised MDE methods while providing additional insight into how different perceptual cues influence depth estimation. These results indicate that perceptual decomposition can support interpretability without sacrificing depth estimation accuracy.

## 1. INTRODUCTION

Reliable depth estimation is vital for many real-world systems, including autonomous driving, robotics, and biomedical applications [1], [2]. To this end, various depth estimation methods have been proposed, many of them based on binocular/stereo vision and active sensing, e.g., time-of-flight (ToF) measurements and laser imaging detection and ranging (LIDAR). Predicting depth with a single camera, known as monocular depth estimation (MDE), is a lower-cost alternative to other methods, with recent advances indicating its increasing competitiveness [3].

Recent advances in deep learning (DL) have enabled self-supervised MDE approaches that exploit geometric and photometric relationships between consecutive frames, eliminating the need for training sets with ground truth depth information [5] - [7]. Although these methods are becoming more effective, their internal decision processes remain largely opaque, in the sense that they produce depth maps without explaining which image features influence the prediction, *e.g.*, when depth estimation fails, it remains unclear whether the error arises from missing texture, misleading color cues, or illumination changes. This lack of transparency limits user trust and can become critical in safety-sensitive applications, *e.g.*, vision tools for robot-assisted treatment.

Existing attempts to address this issue rely mainly on post-hoc tools, which can highlight influential image regions but do not explain how specific visual cues contribute to depth prediction [9] - [11]. Perceptually, depth is inherently multi-cue; humans rely on a combination of texture gradients, color relationships, shading, and structural boundaries to infer spatial layout from a single view. In contrast, conventional RGB-based networks entangle these cues within a single latent space. Separating perceptual image components before depth estimation would offer a more transparent alternative, enabling individual visual signals to be analyzed independently.

Motivated by the growing demand for interpretability, especially in high-risk domains, this work introduces a self-supervised framework for perceptually interpretable monocular depth estimation (PIMDE). The proposed approach is inspired by the interpretation mechanism of Additive Convolutional Neural Networks (ACNNs) and extends it to the problem of depth prediction [10], [11]. To the best of our knowledge, PIMDE is the first framework that inherently interprets MDE in accord with human perception.

The proposed model begins by decomposing the RGB input image into a set of perceptual components, referred to as Perceptual Feature Maps (PFMs). Each PFM is designed to represent a meaningful perceptual quantity for inferring depth. These perceptual components are processed by parallel inverse depth estimation branches; the resulting branch-wise predictions are then fused to produce the final inverse depth map. The final prediction is determined by a known combination of the perceptual branches; hence, it is possible to assess how each PFM contributes to the final estimate. The main contributions of the paper are summarized as follows:

(1) We propose a self-supervised perceptually interpretable framework for MDE that decomposes RGB images into perceptual feature maps to separate chromatic, structural, and luminance visual cues inspired by human visual perception.
(2) We estimate depth from each perceptual cue using our proposed framework, which helps understand how each visual cue affects the final depth prediction.
(3) We use a subtraction-based analysis to show how much each perceptual branch influences the final depth prediction, making the model easier to interpret, and validate the proposed framework on the KITTI benchmark.

The remainder of the paper is organized as follows: section 2 details the methodology, section 3 presents the experimental setup along with quantitative and qualitative results, and section 4 concludes the findings of the paper.

Author version of the paper published in the 2026 IEEE International Conference on Image Processing (ICIP). DOI: https://doi.org/10.1109/ICIP61757.2026.11630094

## 2. METHODOLOGY

### 2.1. Framework Overview

The proposed PIMDE framework is schematically illustrated in Figure 1. The model builds upon the Recurrent Multi-Scale Feature Modulation (R-MSFM) mechanism, which enables progressive depth refinement through iterative modulation of multi-scale feature representations [12]. This mechanism is integrated with the parallel multi-branch design of ACNN, resulting in multiple CNN-based depth estimation subnetworks operating in parallel. Each sub-network operates on a distinct perceptual representation of the input image rather than processing the RGB image directly. This design follows the ACNN principle, in which different perceptual interpretations are handled independently and later combined to produce the final prediction. The input image is decomposed into a set of PFMs, each encoding a specific visual component. These PFMs are fed into parallel depth estimation branches that share the same CNN architecture. Each branch produces a complete inverse depth map, referred to as a perceptual inverse depth estimate (PIDE). The branch-wise PIDEs are fused to obtain the final inverse depth prediction. The inverse depth is defined as the reciprocal of metric depth, i.e., $d^{-1} = 1/d$, which is commonly used in self-supervised monocular depth estimation for stable and efficient optimization [6], [12]. For numerical evaluation, the predicted inverse depth is converted back to metric depth.

### 2.2. Perceptual Feature Map Decomposition

In MDE, depth is inferred from a single image by exploiting visual components that encode scene geometry and spatial layout [13]. Color contrasts, illumination variations, and structural patterns provide complementary information about surface orientation, object boundaries, and relative depth ordering in a scene. Decomposing the input image into PFMs allows these visual cues to be represented explicitly and processed independently, enabling the model to associate different perceptual components with their respective contributions to depth inference. However, these PFMs are not intended to represent complete geometric depth cues: instead, they provide separate perceptual components that can be analyzed to assess their effect on the final depth prediction.

This work adopts an opponent-based perceptual representation of the input image, focusing on color and texture, motivated by the role of color and texture components in human shape and scene perception [14]. Opponent color spaces address limitations of the RGB representation, such as strong channel correlation and poor perceptual alignment. Among them, the perceptually uniform CIE-Lab color space has been widely used in computer vision. CIE-Lab is employed due to the approximate orthogonality of its components. The a and b channels encode red–green (RG) and blue–yellow (BY) chromatic opponency, respectively, while the L channel represents perceptual lightness and contains rich structural and textural information [15]. To further capture texture, the luminance component is analyzed using the 2D Discrete Wavelet Transform (2D-DWT), which provides a multi-resolution representation consistent with human visual perception [16].

The low-frequency approximation of the luminance signal represents light-dark (LD) antagonism, while the highest-frequency band of the first level of the 2D-DWT encodes coarse-fine (CF) texture, which is closely related to edge density and fine structural detail. Consequently, the input to the model is formed as four PFMs, corresponding to RG, BY, CF, and LD, as illustrated in Figure 2.

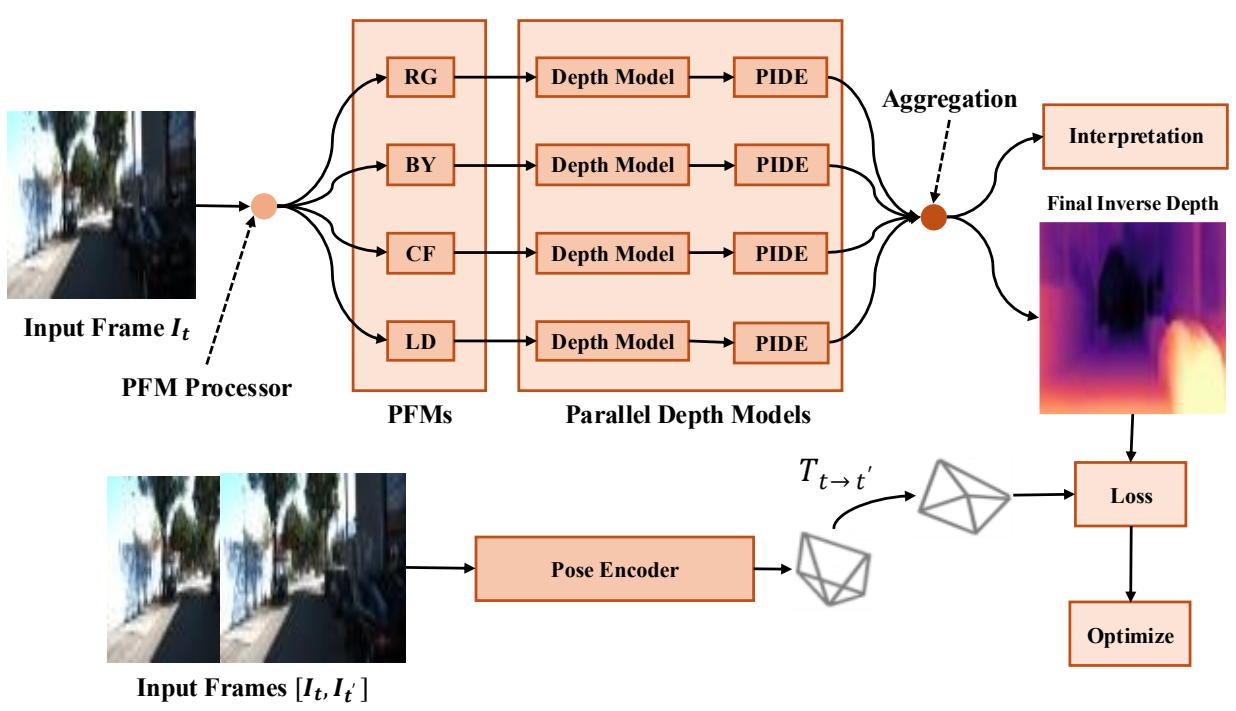

**Figure 1:** Outline of the PIMDE framework

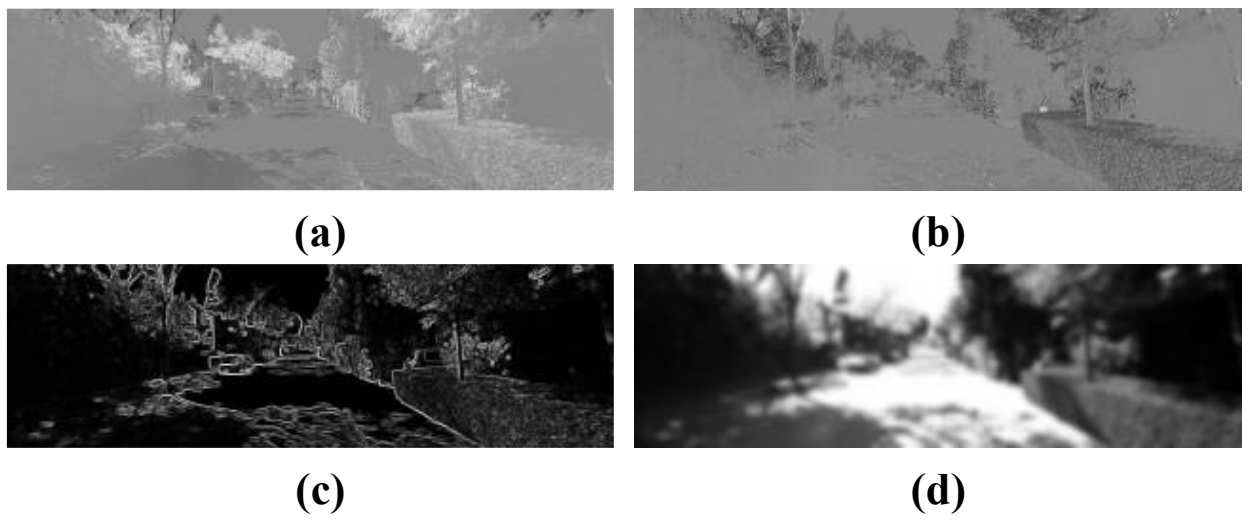

**Figure 2:** Visual representation of PFMs. (a) blue-yellow, (b) red-green, (c) coarse-fine, and (d) light-dark.

### 2.3. Monocular Depth Estimation Model

Given the set of PFMs obtained from the PFM decomposition, MDE is performed using a set of parallel perceptual branches. Each branch is responsible for estimating inverse depth from a single image component, enabling explicit analysis of how individual visual components contribute to depth inference. All perceptual branches employ an identical CNN architecture; this design reflects a core ACNN interpretability framework. So, the variations in the resulting depth estimates are primarily due to differences in perceptual inputs rather than architectural discrepancies.

Let $P = \{P_1, P_2, \dots, P_N\}$ denote the set of PFMs, where each $P_i \in \mathbb{R}^{H \times W}$ corresponds to a specific perceptual component derived from the input image. For each perceptual input $P_i$, a dedicated subnetwork predicts PIDE through a recurrent refinement process, in which an initial inverse depth prediction is repeatedly updated over multiple iterations $s$; each iteration corresponds to one refinement step that improves the prediction by incorporating multi-scale information, as introduced in R-MSFM [12]. The branch-wise PIDE at iteration $s$ is defined as:

$$\hat{d}_i^{(s)} = f_\theta(P_i) \qquad i = 1, \dots, N \tag{1}$$

where $f_\theta(\cdot)$ denotes a convolutional depth estimation network. All branches use the same architecture, while their inputs are different PFMs. Each $\hat{d}_i^{(s)}$ constitutes a PIDE, representing a valid inverse depth map derived solely from the corresponding image component. Importantly, these PIDEs are depth approximations rather than intermediate feature maps, which is essential for enabling branch-level interpretability.

At each iteration s, the final inverse depth is obtained through an explicit aggregation of the branch-wise PIDEs computed as:

$$\hat{d}^{(s)} = \sum_{i=1}^{N} \tilde{\omega}_i \, \hat{d}_i^{(s)} \tag{2}$$

where $\tilde{\omega}_i$ denotes a non-negative normalized fusion weight associated with the $i$-th perceptual branch, satisfying $\sum_i \widetilde{\omega}_i = 1$.

The fusion weights $\widetilde{\omega}_i$ are estimated from branch-influence statistics obtained through a subtraction-based analysis. This procedure quantifies how strongly each perceptual branch influences the final prediction, enabling fusion weights to reflect empirical contribution rather than arbitrary design choices. Specifically, in the first stage, an initial model configuration with uniform weights $\boldsymbol{\omega} = I^N$ is used to generate the fused inverse depth prediction. Then, the influence of each perceptual branch is quantified by measuring the change in the final fused depth map when that branch is removed from the model. From this analysis, the relative difference metric is computed for each branch, providing a scalar measure of the contribution of each branch to the final prediction.

Let $\boldsymbol{\omega} = [\omega_1, \omega_2, \ldots, \omega_N]$ denote the vector of relative difference values across branches. The final fusion weights are obtained via softmax normalization, ensuring non-negativity and unit-sum constraints as:

$$\widetilde{\boldsymbol{\omega}} = softmax(\boldsymbol{\omega}) \tag{3}$$

These weights are computed once from the subtraction-based analysis and are kept fixed during the final training. Thus, the overall fusion weights process follows a two-stage strategy: the model is first trained with uniform weights, and this trained model is then used to estimate the influence of each perceptual branch through subtraction-based analysis. The estimated branch influences are then normalized using Eq. (3) to obtain final fixed fusion weights.

The model is trained in a fully self-supervised manner using monocular image sequences. During training, a pose estimation network predicts the relative camera transformation $T_{t\to s}$ between a target frame $I_t$ and adjacent source frames $I_{t'}$, where $t' \in (t-1, t+1)$, defined as:

$$T_{t\to t'} = PoseNet(I_t, I_{t'}) \tag{4}$$

Given the predicted inverse depth at refinement iteration $s$, a synthesized target view is generated via differentiable view synthesis as given by:

$$I^s_{t'\to t} = I_{t'}(proj(T_{t\to t'}, D^s_t, K)) \tag{5}$$

where $K$ denotes the camera intrinsic matrix, which is identical for all images, $proj(\cdot)$ denotes the projection operator that maps the 3D points obtained from the inverse depth map $D^s_t$ at iteration $s$, after transformation by $T_{t\to t'}$, onto 2D pixel coordinates in the source image $I_{t'}$, using the pinhole camera model [12].

To keep our training process robust against occlusion, the masked photometric reprojection loss is used, following the self-supervised view-synthesis formulation and auto-masking strategy commonly used in MDE [3], formulated as follows:

$$L_p = \sum_{s=1}^{N} \beta^{N-s} \cdot \min_{t'} \mu\left(I_t, I_{t'}, I^s_{t'\to t}\right) \odot pe\left(I_t, I^s_{t'\to t}\right) \tag{6}$$

where $\beta$ emphasizes later refinements and $\odot$ denotes element-wise multiplication. The auto-mask $\mu(\cdot)$ removes pixels violating the motion as defined in (7). $pe(\cdot)$ denotes the minimum per-pixel photometric reprojection loss using SmoothL1 and the structural similarity index measure (SSIM), defined as follows:

$$\mu = \left[\min_{t'} pe\left(I_t, I^s_{t'\to t}\right) < \min_{t'} pe\left(I_t, I_{t'}\right)\right] \tag{7}$$

$$pe\left(I_t, I^s_{t'\to t}\right) = \frac{\alpha}{2}\left(1 - \text{SSIM}\left(I_t, I^s_{t'\to t}\right)\right) + (1-\alpha) SmoothL1\left(I_t, I^s_{t'\to t}\right) \tag{8}$$

where $\alpha$ is a weighting coefficient. Additionally, an edge-aware smoothness regularizer is applied to the mean-normalized inverse depth, defined as:

$$\text{L}_s = \sum_{s=1}^{N} \beta^{N-s} \left|\partial_x d^{s*}_t\right| e^{-\partial_x I_t} + \left|\partial_y d^{s*}_t\right| e^{-\partial_y I_t} \tag{9}$$

where $d^{s*}_t = d^s_t / \bar{d}^s_t$ at iteration $s$, which prevents the inverse depth from approaching zero. So, the final self-supervised loss is formulated as follows:

$$L_{final} = L_p + \lambda L_s \tag{10}$$

where $\lambda$ controls the smoothness regularizer, which helps smooth depth variations in homogeneous regions while preserving depth changes at image boundaries [3].

In the PIMDE framework, the self-supervised loss in (10) is applied primarily to the final fused inverse depth prediction, which represents the final model output. To ensure that each perceptual branch acquires coherent depth, the same loss formulation is additionally applied as auxiliary supervision to each branch's output. So, the overall training objective, which combines the fused loss and the auxiliary branch losses, is formulated as follows:

$$L = L_{final} + \lambda_b \sum_{i=1}^{N} L^i_{branch} \tag{11}$$

where $L_{final}$ is the self-supervised loss applied to the fused prediction, and $L^i_{branch}$ is the same loss as the final loss, applied to the $i^{th}$ branch prediction. $\lambda_b$ controls the contribution of branch-level supervision.

### 2.4. Interpretability

The PIMDE framework is designed to be interpretable, as depth estimation is performed through multiple perceptual branches with a known and fixed fusion mechanism. Each branch processes a single perceptual feature map and produces a corresponding inverse depth estimate that explicitly contributes to the final fused prediction. Let $D_{full} \in R^{H\times W}$ denote the fused inverse depth prediction obtained using all perceptual branches. To assess the contribution of a specific branch $b$, a second fused prediction $D_{-b} \in R^{H\times W}$ is obtained by removing the branch $b$ while keeping all remaining branches unchanged. The contribution of the branch $b$ is then defined through a subtraction-based formulation as

$$\Delta D = D_{full} - D_{-b} \tag{12}$$

This formulation provides a branch-specific contribution map that quantifies how the presence of the branch $b$ affects the final inverse depth prediction. Since the fusion operation is fixed and identical in both cases, the differences between $D_{\text{full}}$ and $D_{-b}$ can be attributed directly to the contribution of the removed perceptual branch. This subtraction-based analysis forms the methodological basis for the interpretability results presented in the subsequent section.

## 3. EXPERIMENTS AND RESULTS

### 3.1. Experimental Setup and Dataset

All experiments were conducted on the KITTI dataset [17], using the standard Eigen split [18]. Following widespread practice in self-supervised monocular depth estimation, static image sequences are removed from the training set to ensure valid photometric supervision [19]. After filtering, the training data consists of 39,810, while the validation set contains 4,424 sequences. The resolution of the input/output is resized to $640 \times 192$ by default.

**Table 1:** Quantitative comparison on the KITTI benchmark dataset. The best and second results for each metric are shown in **bold** and underlined, respectively.

| Models | Train | Depth Errors | | | | Depth Accuracies | | |
|---|---|---|---|---|---|---|---|---|
| | | AbsRel ↓ | SqRel ↓ | RMSE ↓ | RMSE-log ↓ | $\delta < 1.25$ ↑ | $\delta < 1.25^2$ ↑ | $\delta < 1.25^3$ ↑ |
| DF-Net [20] | M | 0.150 | 1.124 | 5.507 | 0.223 | 0.806 | 0.933 | 0.973 |
| Ranjan [21] | M | 0.148 | 1.149 | 5.464 | 0.226 | 0.815 | 0.935 | 0.973 |
| Monodepth [22] | M | 0.148 | 1.344 | 5.972 | 0.216 | 0.816 | 0.941 | 0.976 |
| Monodepth2 [3] | M | **0.115** | 0.903 | 4.863 | 0.193 | **0.877** | **0.959** | **0.981** |
| EPC++ [23] | M | 0.141 | 1.029 | 5.350 | 0.216 | 0.816 | 0.941 | 0.976 |
| Struct2depth [24] | M | 0.141 | 1.026 | 5.291 | 0.215 | 0.816 | 0.945 | 0.979 |
| SGDepth [25] | M | <u>0.117</u> | 0.907 | 4.844 | 0.196 | <u>0.875</u> | <u>0.958</u> | 0.980 |
| **PIMDE-3** (Our) | M | 0.123 | <u>0.831</u> | <u>2.418</u> | <u>0.126</u> | 0.853 | 0.952 | 0.980 |
| **PIMDE-6** (Our) | M | 0.121 | **0.841** | **2.408** | **0.124** | 0.857 | 0.956 | **0.981** |

↓ indicates lower is better, and ↑ indicates higher is better. PIMDE-3/6 denotes the proposed model with 3/6 refinement iterations, respectively. M means the model is trained with self-supervised mono supervision with input resolution 640×192.

Additionally, the smoothness regularization term $\lambda$, the update term $\beta$, the weighting coefficient $\alpha$, and the branch contribution $\lambda_b$ are set to 0.001, 0.9, 0.85, and 0.1, respectively [12]. The model is trained in a fully self-supervised manner using monocular video sequences, without access to ground-truth depth. Adjacent frames are used together with the estimated camera pose to enforce photometric consistency through differentiable view synthesis. During inference, depth is predicted from a single input image, and per-image median scaling is applied during evaluation.

Performance is evaluated using standard depth estimation metrics proposed in [26], including Absolute Relative Error (Abs Rel), Squared Relative Error (Sq Rel), Root Mean Squared Error (RMSE), RMSE in log space (RMSE- log), and the threshold accuracies, whereas accuracies are formulated as:

$$Accuracies = max\left(\frac{d_{pred}}{d_{gt}}, \frac{d_{gt}}{d_{pred}}\right) = \delta < threshold \quad (13)$$

where $M$ is the total number of pixels, $d_{pred}$ and $d_{gt}$ denotes the predicted and ground truth depth values at pixel *p,* respectively. Additionally, the threshold controls the percentage of correct pixels in the estimated depth, which is taken as $1.25, 1.25^2, 1.25^3$.

## 3.2. Quantitative Results

The quantitative performance of the PIMDE framework is evaluated on the KITTI Eigen split and compared with the baseline self-supervised MDE methods, including DF-Net [20], Ranjan [21], Monodepth [22], Monodepth2 [3], EPC++ [23], Struct2depth [24], and SGDepth [25]. Two variants of the proposed approach are considered, corresponding to 3/6 refinement iterations.

As shown in Table 1, both variants achieve competitive performance across all metrics, with the six-iteration model consistently outperforming the three-iteration version. This indicates that additional refinement stages improve depth accuracy while maintaining stable threshold performance. In terms of computational complexity, PIMDE-3 has 14.08M/131.66B, and PIMDE-6 has 15.26M/248.94B parameters/FLOPs, respectively, at the input resolution of $640 \times 192$.

To analyze the role of perceptual cues, single-branch models are trained using one PFM at a time. The full PIMDE configuration, reported in the last row of Table 2, achieves the best overall performance. While the CF branch performs strongly on its own, the complete model consistently outperforms all single-branch settings, highlighting the benefit of combining complementary perceptual information.

**Table 2:** Quantitative evaluation of single-branch PIMDE-6 models trained in individual PFMs. The best and second results for each metric are shown in **bold** and underlined, respectively.

| PFMs | | | | Performance Metrics | | |
|---|---|---|---|---|---|---|
| RG | BY | CF | LD | AbsRel ↓ | RMSE ↓ | $\delta < 1.25$ ↑ |
| ✓ | × | × | × | 0.134 | 2.594 | 0.842 |
| × | ✓ | × | × | 0.128 | 2.512 | 0.847 |
| × | × | ✓ | × | <u>0.126</u> | <u>2.482</u> | <u>0.854</u> |
| × | × | × | ✓ | 0.139 | 2.754 | 0.828 |
| ✓ | ✓ | ✓ | ✓ | **0.121** | **2.408** | **0.857** |

## 3.3. Qualitative Results

Qualitative results provide additional insight into the behavior of PIMDE and complement the quantitative analysis by illustrating how different perceptual cues influence depth estimation. Figure 3 compares qualitative results obtained with three and six refinement iterations. The six-iteration model produces smooth and spatially consistent depth maps, particularly in planar regions, indicating that successive refinement stages progressively correct coarse predictions.

Moreover, Figure 4 visualizes branch-wise PIDEs alongside the fused prediction for the road scene. The CF PIDE produces clearer depth changes along the wall and building in the image. The RG and BY PIDE mainly affect the road surface, where depth changes appear smoother. The LD branch gives the smoothest depth map, especially over the road and distant background, but with fewer visible details. The fused result combines these effects, keeping clear depth changes at scene structures while maintaining smooth depth variation across the image.

## 3.4. Interpretability Analysis

To better understand how different perceptual cues influence the final depth prediction, branch contributions are analyzed using the subtraction-based analysis discussed in the methodology section. Figure 4 illustrates the final depth map, PIDE from each branch, and subtraction-based maps. Each $\Delta D$ map visualizes the change in the fused depth prediction when a specific branch is removed. These maps do not represent depth; instead, they highlight where and how strongly each branch contributes to the final prediction, where lighter colors (e.g., white) indicate smaller differences and darker colors (e.g., red) indicate larger differences

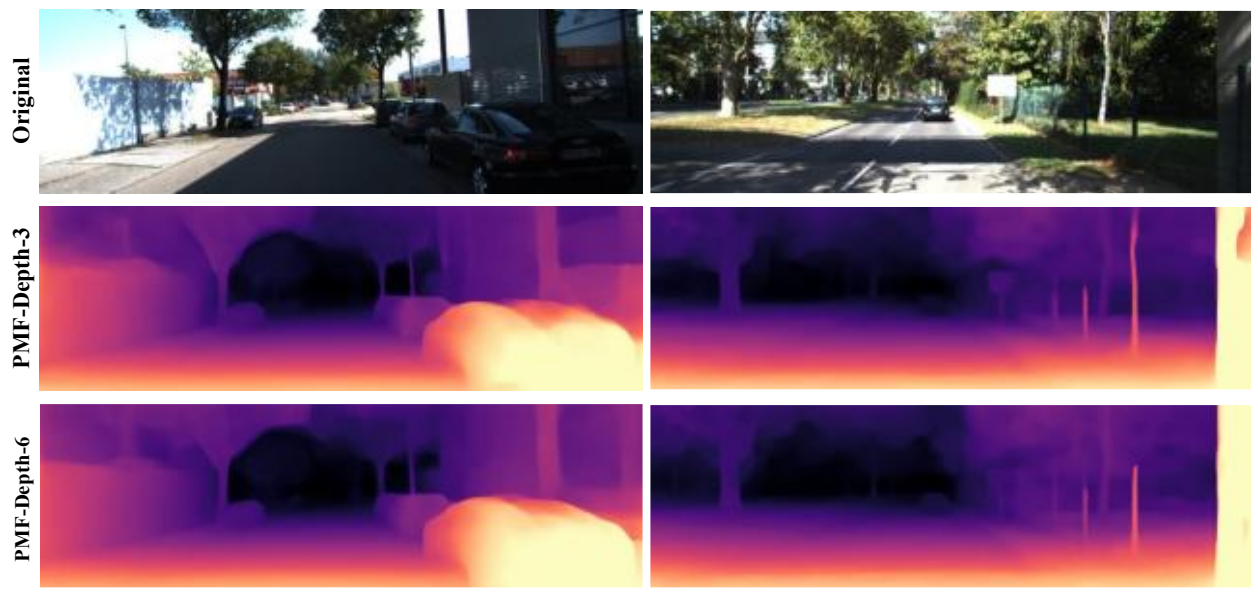

**Figure 3:** Qualitative comparison of PIMDE predictions showing the effect of refinement iterations

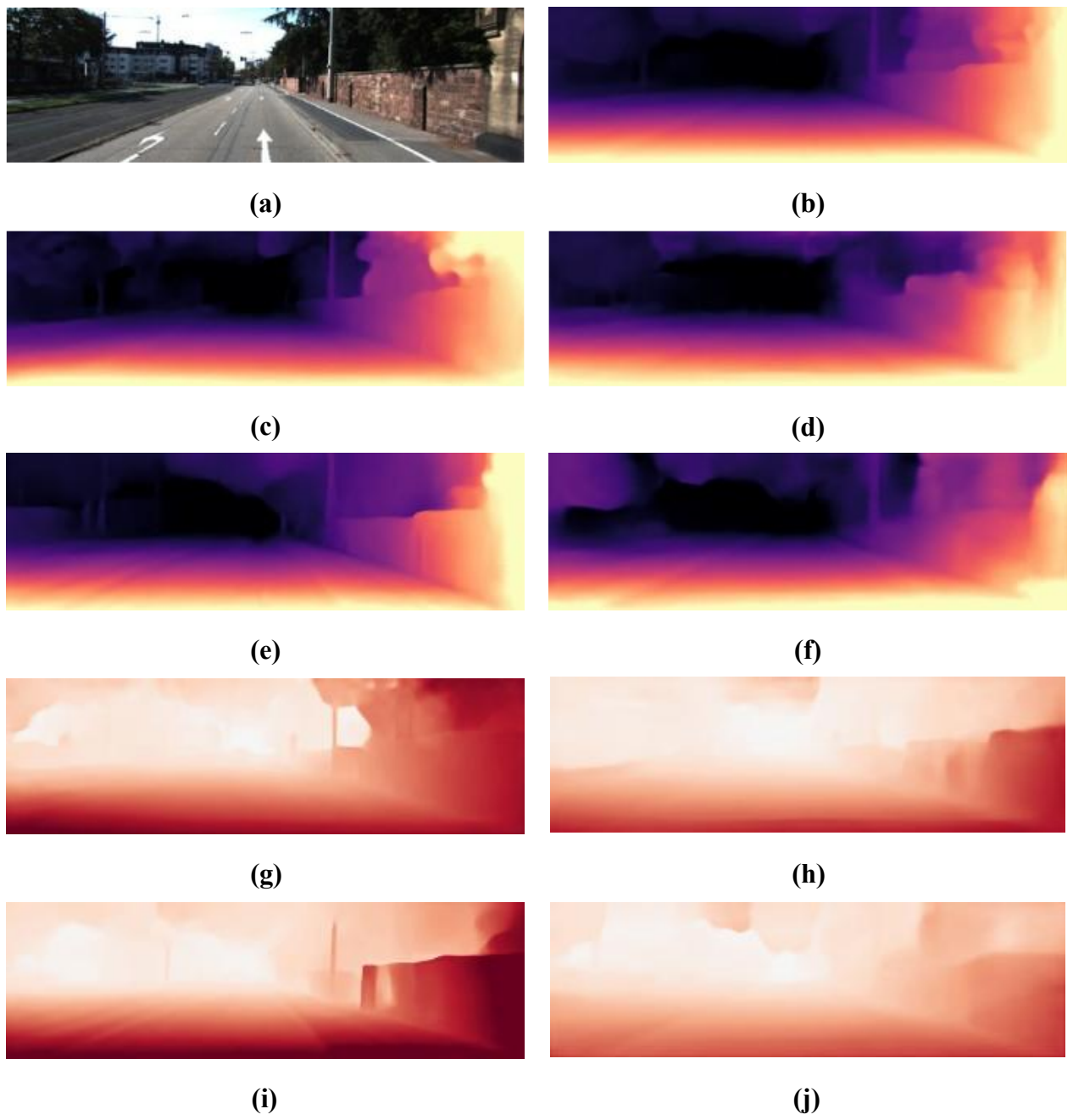

**Figure 4:** Visualization of the final fused inverse depth prediction, branch-wise perceptual inverse depth estimates (PIDEs), and $\Delta D$. (a) Input RGB image, (b) final fused inverse depth map, (c) - (f) PIDEs produced by the BY, RG, CF, and LD branches, respectively, and (g) - (j) $\Delta D$ obtained by removing the BY, RG, CF, and LD branches from the fusion, respectively, where lighter and darker colors indicate smaller and larger differences in the fused prediction, respectively.

The $\Delta D$ maps show that removing the CF branch produces the largest changes (appearing as darker red regions in the respective $\Delta D$), particularly around object boundaries, while removing the LD branch mainly affects broader, smoother regions. The chromatic branches (RG and BY) introduce more localized changes. These observations are consistent with the quantitative branch influence statistics reported in Table 3, where the CF branch exhibits the highest relative difference values, confirming its stronger contribution to the final prediction. This analysis can also be validated from Table 2, where the CF branch has better performance than all other individual branches. Therefore, the proposed framework provides a practical diagnostic view of the model by showing whether the final depth prediction is mainly influenced by structural, luminance, or chromatic information, rather than only visualizing the final depth map.

**Table 3:** Quantitative branch contribution analysis using subtraction-based evaluation.

| Branch | MAD | RMSD | RelDiff |
|---|---|---|---|
| RG | 1.401 ± 0.344 | 1.862 ± 0.557 | 0.225 ± 0.018 |
| BY | 2.295 ± 0.676 | 3.081 ± 1.101 | 0.352 ± 0.036 |
| CF | 2.242 ± 0.452 | 2.896 ± 0.638 | 0.379 ± 0.047 |
| LD | 1.683 ± 0.457 | 2.389 ± 0.244 | 0.244 ± 0.026 |

Branch column indicates when a certain branch is removed during $\Delta D$. MAD: mean absolute difference, RMSD: root mean square difference, RelDiff: relative difference

## 4. CONCLUSION

This work presented PIMDE, a self-supervised MDE framework that introduces PFM decomposition to improve interpretability. By separating chromatic, luminance, and structural cues and processing them through parallel depth estimation branches with a fixed fusion mechanism, the proposed approach allows depth predictions to be examined at the individual perceptual components rather than as black-box outputs. Experimental results on the KITTI dataset show that the PIMDE framework achieves competitive performance compared to various baselines, while providing a clearer interpretation of how different visual components influence the final prediction. Future work will explore additional and automatic selection of PFMs, branch-combination analysis, stronger and domain-specific MDE backbones, and extend the evaluation to additional datasets and domains to further assess the generalization and interpretability of the proposed framework.

## 5. ACKNOWLEDGEMENTS

This project has received funding from the European Union's HORIZON TMA MSCA Doctoral Networks (HORIZON-MSCA-2023-DN-01) under Grant Agreement n°101169012 (Intelli-Ingest Project https://www.intelli-ingest.com/).